\documentclass[11pt,a4paper]{article}
\usepackage[hyperref]{emnlp2020}
\usepackage{times}
\usepackage{latexsym}
\usepackage{graphicx}
\usepackage{float}
\usepackage{import}
\usepackage{amsmath}
\usepackage{booktabs}
\usepackage{amssymb}
\usepackage{url}
\usepackage{multirow}
\usepackage{caption, subcaption}

\usepackage{microtype}

\aclfinalcopy

\title{End-to-End Neural Event Coreference Resolution}

\author{Yaojie Lu${}^{1,3}$, Hongyu Lin${}^{1}$, Jialong Tang${}^{1,3}$, Xianpei Han${}^{1,2}$ , Le Sun${}^{1,2}$ \\
${}^{1}$Chinese Information Processing Laboratory ~ ${}^{2}$State Key Laboratory of Computer Science \\
Institute of Software, Chinese Academy of Sciences, Beijing, China\\
${}^{3}$University of Chinese Academy of Sciences, Beijing, China \\
 {\tt \{yaojie2017,hongyu2016,jialong2019,xianpei,sunle\}@iscas.ac.cn}
}

\date{}

\begin{document}
\maketitle
\begin{abstract}
  Traditional event coreference systems usually rely on pipeline framework and hand-crafted features, which often face error propagation problem and have poor generalization ability.
  In this paper, we propose an \textbf{E}nd-to-\textbf{E}nd \textbf{E}vent \textbf{C}oreference approach -- $\text{E}^{3}\text{C}$ neural network, which can jointly model event detection and event coreference resolution tasks, and learn to extract features from raw text automatically.
  Furthermore, because event mentions are highly diversified and event coreference is intricately governed by long-distance, semantic-dependent decisions, a type-guided event coreference mechanism is further proposed in our $\text{E}^{3}\text{C}$ neural network.
  Experiments show that our method achieves new state-of-the-art performance on two standard datasets.
\end{abstract}

\section{Introduction} \label{sec:introduction}
\begin{figure}[!tpb] 
  \centering
  \setlength{\belowcaptionskip}{-0.5cm}
  \includegraphics[width=0.42\textwidth]{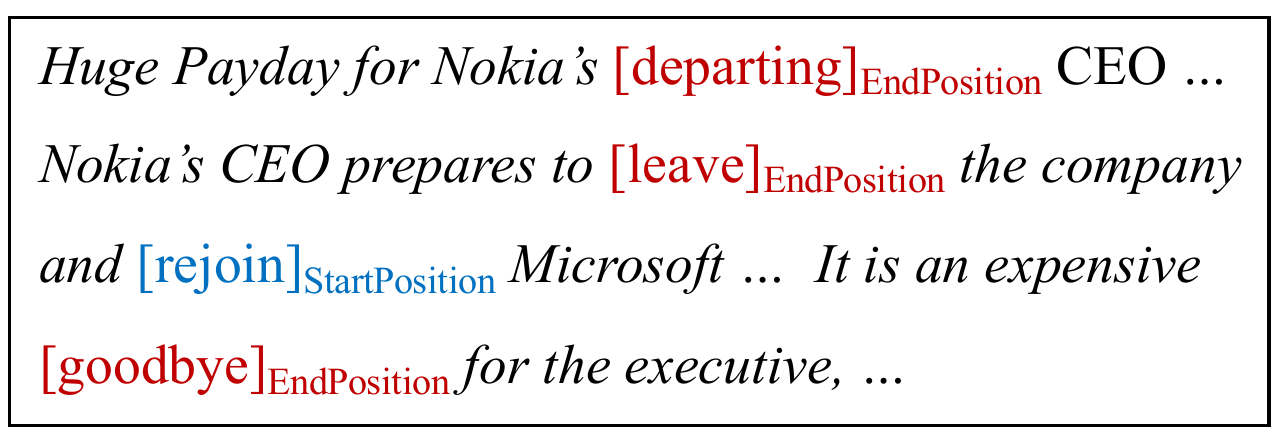}
  \caption{An example of event coreference resolution, which contains two coreferential chains: An \textit{EndPosition} event chain \{\textit{departing}, \textit{leave}, \textit{goodbye}\} and a \textit{StartPosition} chain \{\textit{rejoin}\}.
  }
  \label{fig:example_event_coref}
\end{figure}

Event coreference resolution aims to identify which event mentions in a document refer to the same event \citep{ahn-2006-stages,hovy-etal-2013-events}.
For example, the two event mentions in Figure \ref{fig:example_event_coref}, \textit{departing} and \textit{leave}, refer to the same \textit{EndPosition} event of Nokia's CEO.

Traditional event coreference resolution methods usually rely on a series of upstream components \citep{lu-ng:2018:ijcai2018}, such as entity recognition and event detection.
Such a pipeline framework, unfortunately, often suffers from the error propagation problem.
For instance, the best event detection system in KBP 2017 only achieved 56 F1 \citep{jiang-etal:2017:tac2017}, and it will undoubtedly limit the performance of the follow-up event coreference task (35 Avg F1 on KBP 2017).
Furthermore, most previous approaches use hand-crafted features \citep{chen-etal-2011-unified,lu-ng-joint:2017:acl2017}, which heavily depend on other NLP components (e.g., POS tagging, NER, syntactic parsing, etc.) and thus are hard to generalize to new languages/domains/datasets.

In this paper, we propose an \textbf{E}nd-to-\textbf{E}nd \textbf{E}vent \textbf{C}oreference method -- $\text{E}^{3}\text{C}$ neural network, which can predict event chains from a raw text in an end-to-end manner.
For example, taking the raw text in Figure \ref{fig:example_event_coref} as input, $\text{E}^{3}\text{C}$ will directly output two event coreference chains, \{\textit{departing}, \textit{leave}, \textit{goodbye}\} and \{\textit{rejoin}\}.
By jointly modeling event detection and event coreference, $\text{E}^{3}\text{C}$ neural network does not require any prior components, and the representations/pieces of evidence between different tasks and different decisions can be shared and reinforced.
Besides, $\text{E}^{3}\text{C}$ are learned in an end-to-end manner, which can inherently resolve the error propagation problem.

End-to-end event coreference, however, is challenging due to the mention diversity and the long-distance coreference.
\textit{First, event mentions are highly diversified} \citep{humphreys-etal-1997-event,chen-ji:2009:graph2009}, which may be a variety of syntactic objects, including nouns, verbs, and even adjectives.
For example, an \textit{EndPosition} event can be triggered by \textit{departing}(noun), \textit{leave}(verb), \textit{goodbye}(noun) and \textit{former}(adj).
By contrast, mentions in entity coreference are mostly noun phrases \citep{lu-ng:2018:ijcai2018}.
\textit{Second, coreferential event mentions commonly appear over long-distance sentences, therefore event coreference is intricately governed by long-distance, semantic-dependent decisions} \citep{choubey-huang:2018:acl2018,goyal-etal-2013-structured,peng-etal-2016-event}.
For example, in Figure \ref{fig:example_event_coref} the closest antecedent\footnote{In this paper, antecedents are coreferential mentions that appear earlier in the document.} of the mention \textit{goodbye} -- \textit{leave}, is far from it.
To resolve the coreference between these two distant, diverse event mentions, a system can only rely on their semantic meanings, i.e., they both describe the same \textit{EndPosition} event(the departing of Nokia's CEO) but from different perspectives.
By contrast, most of entity mentions' closest antecedents are in the same or immediately preceding sentence \citep{choubey-huang:2018:acl2018}, which can be resolved more easily using local and syntactic clues.

To resolve the mention diversity problem and the long-distance coreference problem, this paper further proposes a type-guided mechanism into our $\text{E}^{3}\text{C}$ neural network.
This mechanism bridges distant, diverse event mentions by exploiting event type information in three folds:
1) \textbf{type-informed antecedent network} which enables $\text{E}^{3}\text{C}$ to capture more semantic information of event mentions by predicting coreferential scores and type scores simultaneously;
2) \textbf{type-refined mention representation} which enhances mention representation with type information, therefore even lexically dissimilar mentions can be bridged together, such as the two diverse \textit{EndPosition} mentions \textit{goodbye} and \textit{departing};
3) \textbf{type-guided decoding algorithm} which can exploit global type consistency for more accurate event chains.

The main contributions of this paper are:

1. We propose an end-to-end neural network for event coreference resolution –- $\text{E}^{3}\text{C}$ neural network.
$\text{E}^{3}\text{C}$ can jointly model event detection and event coreference, and learn to automatically extract features from raw text.
To the best of our knowledge, this is the first end-to-end neural event coreference model that can achieve state-of-the-art performance.

2. We design a type-guided mechanism for event coreference, which can effectively resolve the mention diversity problem and the long-distance coreference problem in event coreference resolution.

3. We conduct experiments on two standard datasets: KBP 2016 and KBP 2017, which show that $\text{E}^{3}\text{C}$ achieves new state-of-the-art performance.
And additional ablation experiments verify the effectiveness of the proposed type-guided mechanism.

\section{$\text{E}^{3}\text{C}$: End-to-end Neural Event Coreference Resolution} \label{sec:end-to-end-neural-event-coreference}
\begin{figure}[!tpb] 
  \centering
  \includegraphics[width=0.46\textwidth]{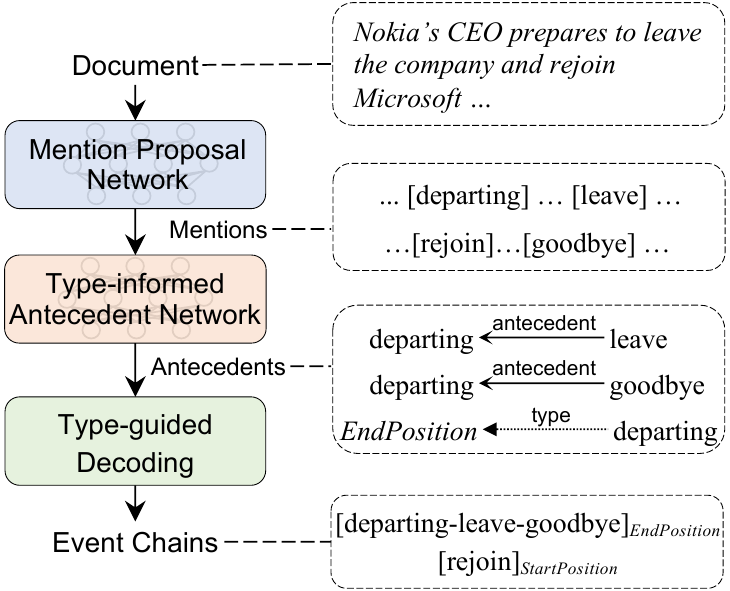}
  \caption{The framework of our $\text{E}^{3}\text{C}$ neural network.}
  \label{fig:framework}
\end{figure}

Given a document $D=\{w_{1}, ..., w_{n}\}$, an end-to-end event coreference system needs to: 1) detect event mentions $\{m_{1}, …., m_{l}\}$ (event detection); 2) predict all coreference chains $\{ev_{*}\}$ (event coreference resolution).
For example, in Figure \ref{fig:example_event_coref}, the mentions are \{\textit{departing}, ..., \textit{goodbye}\} and two coreference chains will be predicted: \{\textit{departing}, \textit{leave}, \textit{goodbye}\}, and \{\textit{rejoin}\}.

To this end, our $\text{E}^{3}\text{C}$ method first detects mentions candidates via a mention proposal network, then identifies all mentions' antecedents via an antecedent prediction network.
To resolve the mention diversity problem and the long-distance coreference problem, a type-guided event coreference mechanism is designed for $\text{E}^{3}\text{C}$.
Figure \ref{fig:framework} shows the framework of our method.
All components in $\text{E}^{3}\text{C}$ are differentiable and can be trained in an end-to-end manner.
In the following, we describe them in detail.

\subsection{Proposing Mention Candidates via Mention Proposal Network}
The mention proposal network detects all event mentions in a document, e.g., identifying \{\textit{departing}, ..., \textit{rejoin}\} as event mentions in Figure \ref{fig:example_event_coref}.
Because event mentions are highly diversified expressions (e.g., \textit{goodbye}, \textit{former} and \textit{leave} for \textit{EndPosition}), we first capture the semantic information of all tokens via a contextualized representation layer, then identify mention candidates via a mention proposal layer.
The details are as follows.

\paragraph{Contextualized Word Representation Layer.}
To capture the semantic information for proposing event mentions, we learn a contextualized representation for each token.
Concretely, we first obtain a task-independent representation for each token based on pre-trained BERT embeddings \citep{devlin-etal:2019:bert}.
Following \citet{tenney-etal:2019:iclr2019}, a token $w_{i}$'s representation $\mathbf{h}_{i}\in \mathbb{R}^{d}$ is pooled across different BERT layers using scalar mixing \citep{peters-etal:2018:naacl2018} as $\mathbf{h}_{i} = \gamma \sum_{j=1}^{L} \alpha_{j} \mathbf{x}_{i}^{(j)}$,
where $\mathbf{x}_{i}^{(j)}$ is the embedding of token $i$ from BERT layer $j$, $d$ is size of bert embedding, $\alpha_{j}$ is softmax-normalized weights, and $\gamma$ is a scalar parameter.

Because event arguments can provide critical evidence \citep{bejan-harabagiu:2010:ACL,Lee:2012:JEE:2390948.2391006,mcconky2012improving,cybulska-vossen-2013-semantic}, we further obtain an event-specific token representation by distilling argument information from raw text implicitly.
Specifically, we design a mask attention strategy \citep{dong-etal:2019:nips2019}.
Given task-independent token representations $\mathbf{H}=\{\mathbf{h}_1, \mathbf{h}_2,..., \mathbf{h}_{n}\}$, our attention mechanism first models the relevance between tokens via a scaled dot-product attention \citep{Ashish-etal:2017:nips2017} without linear projection, and then computes the final contextualized word representations $\mathbf{C}=\{\mathbf{c}_1, \mathbf{c}_2,..., \mathbf{c}_{n}\}$ as:
\begin{equation}
  \setlength{\abovedisplayskip}{3pt}
  \setlength{\belowdisplayskip}{3pt}
  \begin{gathered}
  \mathbf{C} = \mbox{softmax}(\frac{\mathbf{H}\mathbf{H}^{T}}{\sqrt{d}} + \mathbf{M}) \mathbf{H}\\
  \mathbf{M}_{ij} = \left\{
    \begin{array}{ll}
    0, & |i - j| < c \\
    -\infty & \text{otherwise} \\
  \end{array}
    \right.\\
  \end{gathered}
\end{equation}
where $c$ is the size of local window (this paper focuses on the local context since arguments empirically appear around event mentions\footnote{In KBP 2017 training set, about 90\% of arguments appear in the $\pm$10-word window of their trigger word.}, and we set $c=10$ in this paper), and $\sqrt{d}$ is the scaling factor.

\paragraph{Mention Proposal Layer.}\label{sec:mention_proposal_layer}
Given the token representations, the mention proposal layer assigns a mention score to each span -- $s_{m}(i)$, which indicates the likelihood for span $i$ being an event mention.
For example, in Figure \ref{fig:example_event_coref} the mention proposal layer will assign spans \{\textit{departing}, \textit{leave}, \textit{rejoin}, ...\} with high $s_{m}(i)$ scores because they are highly likely to be event mentions, and assign spans \{\textit{prepares to}, \textit{company}, ...\} with low $s_{m}(i)$ scores because they are unlikely to be event mentions.

Given all spans within a restricted length\footnote{This paper restricts span length to 1, which can cover 96.6\% mentions in KBP 2017 training set. For this case, the attented span representation $\mathbf{g}_{i}$ is equivalent to $\mathbf{c}_{i}$.} in a document, the mention proposal layer represents each span $i$ as $\mathbf{g}_{i} = \mathbf{\hat{c}}_{i}$, where $\mathbf{\hat{c}}_{i}$ is the soft head attention-based aggregation of all token representations in span $i$ \citep{lee-etal:2016:arxiv}.
Given $\mathbf{g}_{i}$, the mention score $s_{m}(i)$ is computed via standard feed-forward neural networks: 
\begin{equation}
  \setlength{\abovedisplayskip}{3pt}
  \setlength{\belowdisplayskip}{3pt}
  \label{equ:mention_score_layer}
  s_{m}(i) = \mbox{FFNN}_{m}(\mathbf{g}_{i})
\end{equation}
Finally, we rank all spans according to their mention scores \citep{lee-etal:2017:emnlp2017}, and only retain top-$l$ mentions\footnote{In this paper, $l = 0.1 \times$ document length.} $\{m_{1}, m_{2}, ..., m_{l}\}$ as event mention candidates for computation efficiency.

\subsection{Predicting Antecedent via Type-informed Antecedent Network} \label{sec:type-informed-antecedent}
Given an event mention, the type-informed antecedent network predicts its antecedents, and the antecedent predictions can be used as local pair-wise coreference decisions.
For example, our method will predict the antecedent of \textit{leave} as \textit{departing} in Figure \ref{fig:example_event_coref} and $\langle$departing, leave$\rangle$ can be used as a pair-wise coreference decision.

For each mention $m_{i}$ in $\{m_{1}, ..., m_{l}\}$, the type-guided antecedent network produces two kinds of scores simultaneously:
1) $s(i, j)$ -- the score for mention $m_{j}$ being antecedent of $m_{i}$, where $m_{j}$ must appear before $m_{i}$ in the document;
2) $s(i, t_{k})$ -- the score for mention $m_{i}$'s type being $t_{k}$.

\paragraph{Antecedent Score.}
Given a mention $m_{i}$, antecedent network computes an antecedent score $s(i, j)$ for each mention pair $\langle m_{i}, m_{j} \rangle$:
\begin{equation}
  \setlength{\abovedisplayskip}{4pt}
  \setlength{\belowdisplayskip}{4pt}
  s(i, j) = s_{m}(i) + s_{m}(j) + s_{a}(i, j)
\end{equation}
where $j < i$, $s_{m}(i)$ and $s_{m}(j)$ are the mention scores described in $\S$\ref{sec:mention_proposal_layer}; $s_{a}(i, j)$ measures the semantic similarity between $m_{i}$ and $m_{j}$, computed via a standard feed-forward neural network:
\begin{equation}
  s_{a}(i, j) = \mbox{FFNN}_{a}([\mathbf{g}_{i}, \mathbf{g}_{j}, \mathbf{g}_{i} \circ \mathbf{g}_{i}, \Phi(i, j)])
\end{equation}
where $\mathbf{g}_{i} \circ \mathbf{g}_{i}$ is the element-wise similarity of each mention pair $\langle m_{i}, m_{j} \rangle$, and $\Phi(i, j)$ is the distance encoding between two mentions.

\paragraph{Event Type Score.}
As described in $\S$\ref{sec:introduction}, event coreference is intricately governed by long-distance, semantic-dependent decisions.
To address this issue, this paper exploits event type information for better event coreference resolution.
Specifically, besides  antecedent prediction for each mention, we further predict its event type so that: 
1) the neural network will be guided to capture more semantic information about event mentions \citep{durrett-klein-2014-joint};
2) the type information ensures the global type consistency during coreference resolution, i.e., mentions in the same coreference chain will have the same event type.

Specifically, we first embed all event types $\mathcal{T} = \{t_{1}, ..., t_{t}\}$ via a hierarchical embedding algorithm.
The embedding of $t_{k}$ is $\mathbf{g}_{t_{k}} = \mathbf{W}_{e} \cdot [\mathbf{e}_{event}, \mathbf{e}_{type}(t_{k})]$, where $\mathbf{e}_{event}$ is shared by all event types, $\mathbf{e}_{type}(t_{k})$ indicates embedding of $t_{k}$, 
and $\mathbf{W}_{e}$ is a mapping matrix.
The dimension of $\mathbf{g}_{t_{k}}$ is the same as mention embedding $\mathbf{g}_{i}$.

Then the type scores $s(i, t_{k})$ are computed via the same scoring function for antecedent prediction:
\begin{equation}
  \setlength{\abovedisplayskip}{1pt}
  \setlength{\belowdisplayskip}{1pt}
  \begin{gathered}
    s_m({t_{k}}) = \mbox{FFNN}_{m}(\mathbf{g}_{t_{k}}) \\
    s_{a}(i, t_{k}) = \mbox{FFNN}_{a}([\mathbf{g}_{i}, \mathbf{g}_{t_{k}}, \mathbf{g}_{i} \circ \mathbf{g}_{t_{k}}, \Phi(i, t_{k})])\\
    s(i, t_{k}) = s_{m}(i) + s_{m}(t_{k}) + s_{a}(i, t_{k})
  \end{gathered}
\end{equation}
where the distance for $\Phi(i, t_{k})$ is zero in type scores computation.

For non-mention spans, we add a dummy antecedent $\varepsilon$ and assign the antecedents of all non-mention spans to $\varepsilon$, e.g., \textit{company} and \textit{prepares to} in Figure \ref{fig:example_event_coref}.
We fix the score $s(i, \varepsilon)$ to 0, and identify a span $i$ as non-mention span if all its antecedent scores $s(i, j) \leq 0$ and all type scores $s(i, t_{k}) \leq 0$.

In this way, we obtain the antecedent scores and the type scores for each mention via our type-informed antecedent network.

\subsection{Enhancing Mention Representation via Type-based Refining}
In this section, we describe how to further refine a mention’s representation using its type information, so it can capture more semantic information for event coreference resolution.
For example, although \textit{goodbye} and \textit{departing} are lexically dissimilar, we can still capture their semantic similarity by further encoding their event type information, i.e., both of them have the same event type -- \textit{EndPosition}.

To refine mention representation, we first define a probability distribution $Q(t_{k})$ over all event types $\mathcal{T}$ and $\{\varepsilon\}$ for each mention span $m_{i}$:
\begin{equation}
  \setlength{\abovedisplayskip}{3pt}
  \setlength{\belowdisplayskip}{3pt}
  Q(t_{k}) = \frac{e^{s(i, t_{k})}}{\sum_{t_k'\in \mathcal{T} \bigcup \{\varepsilon\}} e^{s(i, t_k')}}
\end{equation}
where $s(i, t_k)$ is the type score.
We then obtain an expected event type representation $\tilde{\mathbf{g}}_{i}$ for each span $m_{i}$ using the type distribution $Q(t_{k})$ as:
\begin{equation}
  \setlength{\abovedisplayskip}{3pt}
  \setlength{\belowdisplayskip}{3pt}
  \tilde{\mathbf{g}}_{i} = \sum_{t_k'\in \mathcal{T} }Q(t_k=t_k') \cdot \mathbf{g}_{t_k'} + Q(t_k=\varepsilon) \cdot \mathbf{g}_{i}
\end{equation}
Then, we obtain a refined span representation $\mathbf{g}'_{i}$ by combining its expected event type representation $\tilde{\mathbf{g}}_{i}$ and its original span representation $\mathbf{g}_{i}$ via a learnable adaptive gate $\mathbf{f}_{i}$:
\begin{equation}
  \setlength{\abovedisplayskip}{3pt}
  \setlength{\belowdisplayskip}{3pt}
  \begin{gathered}
    \mathbf{g}'_{i} = \mathbf{f}_{i} \circ \mathbf{g}_{i} + (1 - \mathbf{f}_{i}) \circ \tilde{\mathbf{g}}_{i}\\
    \mathbf{f}_{i} = \sigma(\mathbf{W}_{f} \cdot [\mathbf{g}_{i}, \tilde{\mathbf{g}}_{i}])
  \end{gathered}
\end{equation}
where $\mathbf{W}_{f}$ is a weight matrix.

Finally, the antecedent network will recompute the coreferential antecedent score $s'(i, j)$ and event type score $s'(i, t_{k})$ using  the refined span representation $\mathbf{g}'_{i}$.

\subsection{Coreference Resolution via Type-guided Decoding}
\label{sec:type-consistent-decoding}
The type-informed antecedent network produces pairwise coreference scores of mention pairs.
To form coreference chains, a naive approach is to directly connect all mentions using their highest-scored antecedent.
Unfortunately, such a greedy decoding algorithm only considers local pair-wise consistencies, their results may not be globally optimal, e.g., a coreference chain may contain mentions with different event types.

To address this issue, we propose a decoding algorithm, which can ensure the global consistency of a coreference chain through a type-guided mechanism.
For example, to resolve the chain \{\textit{departing}, \textit{leave}\} in Figure \ref{fig:example_event_coref}, $\text{E}^{3}\text{C}$ considers both the antecedent score of $\langle$\textit{departing}, \textit{leave}$\rangle$, and the type consistency that both \textit{departing}, \textit{leave} are \textit{EndPosition} mentions.

Concretely, given the mentions \{$m_{1}$, ..., $m_{l}$\} in a document $D$, $\text{E}^{3}\text{C}$ constructs the event coreference chains by sequentially identifying the best antecedent of each mention, further consider the type consistency.
For each mention $m_{i}$, we first find the mention $a_{i}$ which has the max coreferential score with $m_{i}$ where $m_j$ appears before $m_{i}$:
\begin{equation}
  \setlength{\abovedisplayskip}{0pt}
  \setlength{\belowdisplayskip}{0pt}
  a_{i}=\arg\max_{m_{j}, \, j<i} s(i, j)
\end{equation}
and then we check the type consistency between $\langle a_{i}, m_{i} \rangle$ by comparing their antecedent score $s(i, a_{i})$ and the type prediction score of $m_{i}$, $s(i, t_{i})$.
If $s(i, a_{i}) > s(i, t_{i})$, $\text{E}^{3}\text{C}$ considers $m_{i}$ and $a_{i}$ as type consistent and links mention $i$ to $a_{i}$;
otherwise, when $s(i, a_{i}) \leq s(i, t_{i})$, $\text{E}^{3}\text{C}$ considers $m_{i}$ and $a_{i}$ as type inconsistent and starts a new event chain for $m_{i}$ with its type $t_{i}$.

\section{Model Learning} \label{sec:model_learning}
This section describes how to learn $\text{E}^{3}\text{C}$ neural network in an end-to-end manner.
Given a training corpus $\mathcal{D} = \{D_{1}, ..., D_{N}\}$ where each instance $D_{i}$ is a document with its event mention, mention type, and coreference annotations, our objective function contains two parts: $\mathcal{L}_{antecedent}(\mathbf{\Theta})$ -- the antecedent loss, and $\mathcal{L}_{proposal}(\mathbf{\Theta})$ -- the mention proposal loss:
\begin{equation}
  \setlength{\abovedisplayskip}{6pt}
  \setlength{\belowdisplayskip}{6pt}
  \mathcal{L}(\mathbf{\Theta}) = \mathcal{L}_{antecedent}(\mathbf{\Theta}) + \lambda\mathcal{L}_{proposal}(\mathbf{\Theta})
\end{equation}
where $\lambda$ is the coefficient of mention proposal loss (we set $\lambda =1$ in this paper).
This paper optimizes $\mathbf{\Theta}$ by maximizing $\mathcal{L}(\mathbf{\Theta})$ via Adamax \citep{kingma:2015:adam}.
The two losses are as follows.

\begin{figure}[!tpb] 
  \setlength{\belowcaptionskip}{-0.5cm}
  \centering
  \includegraphics[width=0.40\textwidth]{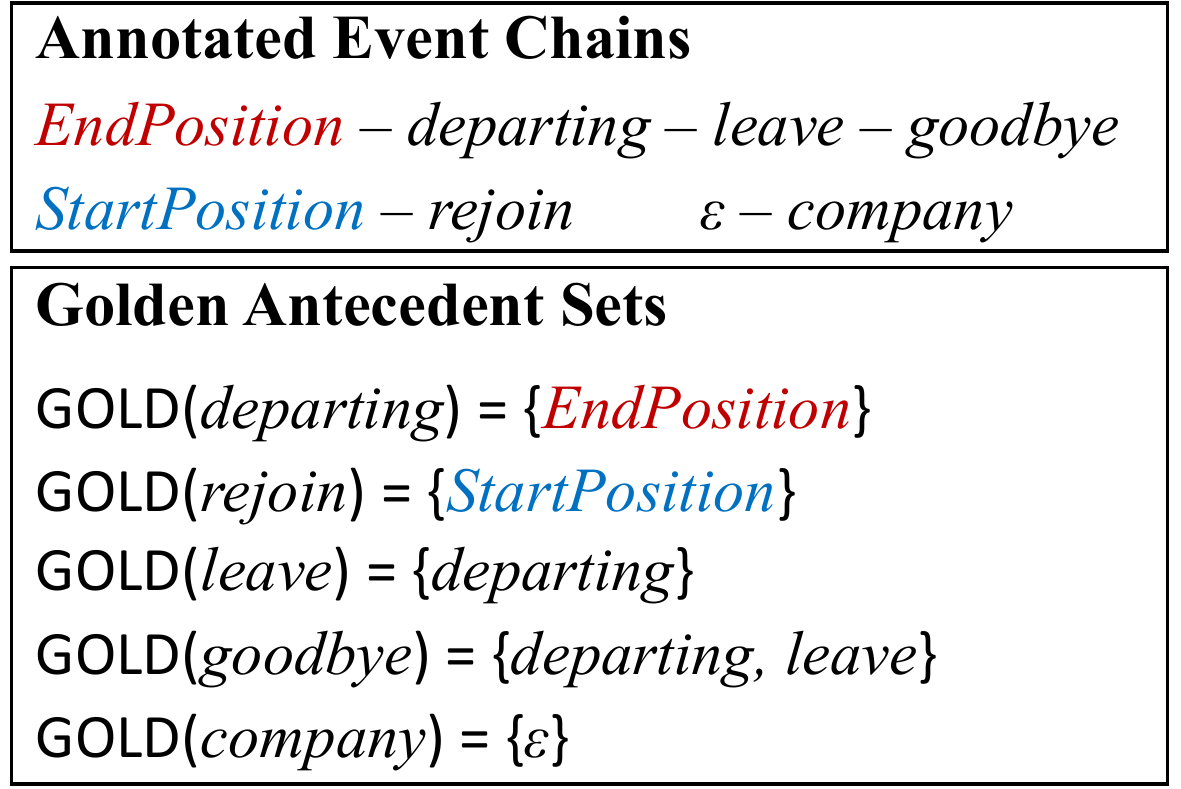}
  \caption{An illustration of gold antecedent sets.
  }
  \label{fig:gold_antecedent_sets}
\end{figure}

\paragraph{Antecedent Loss.}
It measures whether a mention links to its correct antecedent.
For each mention $m_{i}$, this paper identifies its gold antecedent set $\text{GOLD}(i)$ as shown in Figure \ref{fig:gold_antecedent_sets}:

1) For the first mention of an event chain, the gold antecedent is its event type.
For example, the gold antecedent set of \textit{departing} is \{\textit{EndPosition}\}.

2) For remaining mentions in a chain, the gold antecedents are all its coreferential antecedents.
For example, the gold antecedent set of \textit{goodbye} is \{\textit{departing}, \textit{leave}\}.

3) For non-mention spans, the gold antecedent is the dummy antecedent $\varepsilon$. For example, the gold antecedent set of \textit{company} is \{$\varepsilon$\}.

Given $\text{GOLD}(i)$ for each $m_{i}$ in top-$l$ mention set of document $D$, the antecedent loss function is a margin log-likelihood function:
\begin{equation}
  \setlength{\abovedisplayskip}{0pt}
  \setlength{\belowdisplayskip}{0pt}
  \begin{gathered}
    \mathcal{L}(\mathbf{\Theta})_{antecedent} = \log \prod_{i=1}^{l} \sum_{\hat{y}\in \mathcal{Y}(i) \cap \text{GOLD}(i)} P(\hat{y}|D)\\
    P(y_{i}|D) = \frac{\exp(s(i, y_{i}))}{\sum_{y'\in \mathcal{Y}(i)} \exp(s(i, y'))}\\
  \end{gathered}
\end{equation}
where $\mathcal{Y}(i)$ is the valid antecedent set for $m_{i}$.

\paragraph{Mention Proposal Loss.} It measures whether our model can accurately identify event mentions.
Specifically, the mention proposal loss uses the binary cross-entropy loss function of the mention proposal network:
\begin{equation}
  \setlength{\abovedisplayskip}{0pt}
  \setlength{\belowdisplayskip}{0pt}
  \begin{aligned}
    \mathcal{L}(\mathbf{\Theta})_{proposal} = & \sum_{i = 1}^{n}  y_{i} \log \sigma(s_{m}(i)) \\
    & + (1 - y_{i}) \log (1 - \sigma(s_{m}(i))) \\
  \end{aligned}
\end{equation}
where $\sigma$ is the sigmoid function, $y_{i}=1$ indicates span $i$ is an event mention, otherwise $y_{i}=0$.

\section{Experiments} \label{sec:experiments}
\subsection{Datasets} \label{sec:dataset}

\begin{table*}[!t]
    \centering
    \resizebox{0.95\textwidth}{!}{
    \begin{tabular}{clccccc|c}
      \toprule
      % \hline
      & 
      & \multicolumn{1}{c}{\textbf{Type-F1}} 
      & \multicolumn{1}{c}{\textbf{B}$^3$} 
      & \multicolumn{1}{c}{\textbf{CEAF}$_e$} 
      & \multicolumn{1}{c}{\textbf{MUC}} 
      & \multicolumn{1}{c}{\textbf{BLANC}} 
      & \multicolumn{1}{c}{\textbf{AVG-F}} \\
      \midrule
      \multirow{5}{*}{KBP 2016} & Top 1 in TAC 2016 \citep{lu-ng:2016:tac2016} & 46.99 & 37.49 & 34.21 & 26.37 & 22.25 &  30.08 \\
      & Mention Ranking \citep{lu-ng-mention-ranking:2017:icmla2017} & 46.99 & 38.64 & 36.16 & 26.30 & 23.59 & 31.17\\
      & Joint Model \citep{lu-ng-joint:2017:acl2017} & 49.30 & 40.90 & 39.00 & 27.41 & 25.00 & 33.08 \\
      & Interact Model$_{\text{BERT}}$ & 53.65 & 41.71 & 38.75 & 32.17 & 25.90 & 34.63 \\ 
      & E$^3$C (\textit{this paper}) & \textbf{55.38} & \textbf{46.32} & \textbf{45.19} & \textbf{34.39} & \textbf{28.74} & \textbf{38.66} \\
      \midrule 
      \multirow{5}{*}{KBP 2017} & Top 1 in TAC 2017 \citep{jiang-etal:2017:tac2017} & 56.19 & 43.84 & 39.86 & 30.63 & 26.97 & 35.33 \\
      & Interact Model \citep{huang-etal:2019:naacl2019}  & - & 42.84 & 39.01 & 31.12 & 24.99 & 34.49 \\
      & \, \, + Transfer \citep{huang-etal:2019:naacl2019} & - & 43.20 & 40.02 & 35.66 & \textbf{32.43} & 36.75 \\
      & Interact Model$_{\text{BERT}}$ & 56.54 & 45.82 & 44.89 & 33.61 & 28.49 & 38.20   \\
      & E$^3$C (\textit{this paper}) & \textbf{58.33} & \textbf{47.77} & \textbf{45.97} & \textbf{39.06} & 30.60 & \textbf{40.85} \\
      \bottomrule
    \end{tabular}
    }
    \caption{Overall performance on KBP 2016 and KBP 2017 datasets, the results of baselines are adapted from their original papers.}
    \label{tab:overall-performance}
  \end{table*}%

Following previous studies \citep{lu-ng:2016:tac2016,lu-ng-joint:2017:acl2017,jiang-etal:2017:tac2017,huang-etal:2019:naacl2019}, we use KBP 2016 and KBP 2017 English datasets for evaluation\footnote{
  There are also other public event coreference datasets: Ontonotes \citep{Pradhan:2007:UCI:1304608.1306369}, ECB+ \citep{bejan-harabagiu-2008-linguistic,cybulska-vossen:2014:LREC2014}, ACE \citep{ace2005:2005:LDC2005}.
  Ontonotes and ECB+ are not annotated with event type information therefore is not appropriate for evaluating our end-to-end event coreference model.
  ACE dataset has strict notion of event identity \citep{song-etal:2015:eventworkshop,lu-ng-joint:2017:acl2017}, which requires which two event mentions coreferential if and only if ``they had the same agent(s), patient(s), time, and location". 
  Because $\text{E}^{3}\text{C}$ don't perform argument extraction for event coreference, ACE isn't used in this paper.
  For fair comparsion, we choose the KBP datasets \citep{ellis-etal:2015:tac2015,ellis-etal:2016:tac2016,getman-etal:2017:tac2017} so that different systems can be compared in the same settings.
}:

\textbf{KBP 2016.} For KBP 2016, we use the same setup as \citet{lu-ng-joint:2017:acl2017}, i.e., 509 documents for training, 139 documents for parameter tuning, and the official KBP 2016 eval set for evaluation.

\textbf{KBP 2017.} Following \citet{huang-etal:2019:naacl2019}, we use the English portion of KBP 2015 and 2016 dataset for training, and the KBP 2017 dataset for evaluation.
We sample 50 documents from the 2016 evaluation dataset as the validation set.

\subsection{Baselines} \label{sec:baselines}
We compare $\text{E}^{3}\text{C}$ with the following baselines\footnote{Different from the official type-constraint settings in KBP 2016 and KBP 2017, \citet{choubey-huang:2018:acl2018} used relaxed constraints without considering event mention type, so we exclude their system for fair comparison.}:

\textbf{Multi-Pass Sieve} \citep{lu-ng:2016:lrec2016} is an iterative pipeline-based method, which uses both hand-crafted rules and automatic classifiers.
We compare two such systems: 
the Top 1 system in TAC 2016 \citep{lu-ng:2016:tac2016} and the Top 1 system in TAC 2017 \citep{jiang-etal:2017:tac2017}, 
both of which use additional ensemble strategy for better event detection performance.

\textbf{Mention Ranking} \citep{lu-ng-mention-ranking:2017:icmla2017} ranks the candidate antecedents of all event mentions and selects the top-ranked antecedent for each mention.

\textbf{Joint Model} \citep{lu-ng-joint:2017:acl2017} is a hand-crafted feature-based system that addresses the error propagation problem by jointly learning event trigger detection, event coreference resolution, and event anaphoricity prediction tasks.

\textbf{Interact Model} \citep{huang-etal:2019:naacl2019} is the state-of-the-art pair-wise method which decides whether two mentions are coreferential using an interactive binary classifier, and then link coreferential mentions to produce final event chains.
We also compare with an enhanced model that transfers argument compatibility features from external unlabeled data – Interact Model + Transfer.
We also reimplement the interact model using BERT as its feature extractor – Interact Model$_{\text{BERT}}$, therefore E$^{3}$C and Interact Model can be compared with the same feature extractors.

\subsection{Evaluation Metrics} \label{sec:evaluation_metrics}
We use the standard evaluation metrics in KBP evaluation, and compute them using the official evaluation toolkit\footnote{https://github.com/hunterhector/EvmEval}.
We use 4 measures: MUC \citep{vilain-etal:1995:MUC1995}, $\text{B}^3$ \citep{bagga1998:1998:b3}, $\text{CEAF}_{e}$ \citep{luo:2005:EMNLP2005}, and BLANC \citep{recasens2011blanc:2011:NaturalLanguageEngineering}.
Following previous studies \citep{lu-ng-joint:2017:acl2017,huang-etal:2019:naacl2019}, the primary metric AVG-F is the unweighted average of the above four F-scores.
We also report the event detection performance using the typed F1-scores as Type-F1.

\subsection{Overall Performance} \label{sec:overall_performance}
Table \ref{tab:overall-performance} shows the overall performance on KBP 2016 and KBP 2017.
We can see that:

1. \textbf{$\text{E}^{3}\text{C}$ neural network achieves state-of-the-art performance on both datasets.}
Compared with all baselines, $\text{E}^{3}\text{C}$ gains at least 11.6\% and 6.9\% AVG-F improvements on KBP 2016 and KBP 2017, respectively.
This verifies the effectiveness of the end-to-end framework and the type-guided event coreference mechanism.

2. \textbf{By jointly modeling all tasks together and learning all components in an end-to-end manner, $\text{E}^{3}\text{C}$ neural network significantly outperforms pipeline baselines.}
Compared with Interact Model$_{\text{BERT}}$ which uses the same BERT-based feature extractors, $\text{E}^{3}\text{C}$ still gains 3.2\% Type-F1 and 6.9\% AVG-F improvements on KBP 2017.
This verified the effectiveness of the end-to-end training on reducing the error propagation problem.
Besides, by modeling all tasks together, representations and pieces of evidence can be shared and reinforced between different decisions and tasks.

\subsection{Detailed Analysis}\label{sec:detailed_analysis}

In this section, we analyze the effects of type-guided mechanism, end-to-end learning, and pre-trained models.

\paragraph{Effect of Type Guided Mechanism.}
To investigate the effect of type-guided mechanism in $\text{E}^{3}\text{C}$, we conduct ablation experiments by ablating type-refined representation (-Type-Refined) and by replacing type-guided decoding with the naive best antecedent decoding (-Type-Guided).
Type Rule is a simple heuristic method that regards all event mentions in the same type are coreferential.
The results are shown in Table \ref{tab:ablation-analysis-type}.
We can see that:

1) Type-guided decoding is effective for event coreference.
By considering both type consistency and antecedent score, $\text{E}^{3}\text{C}$ obtains an 8.1\% (3.05) AVG-F improvement over naive decoding.

2) Type-refined representation helps resolve the mention diversity problem and the long-distance coreference problem.
By incorporating type information into mention representation, $\text{E}^{3}\text{C}$ obtains a 3.1\% (1.24) AVG-F improvement.

\paragraph{Effect of End-to-end Learning.}
To investigate the effect of end-to-end learning, we conduct experiments on three variations of $\text{E}^{3}\text{C}$:
$\text{E}^{3}\text{C}_\text{Two Stage}$ which models event mention detection and coreferential antecedent prediction in two independent models but they share span embeddings;
$\text{E}^{3}\text{C}_{\text{w/o Proposal Loss}}$ which removes the mention proposal loss;
$\text{E}^{3}\text{C}_\text{GoldMention}$ which uses gold mentions for coreference resolution and type scoring, but the model still needs to predict the type of each mention.
Table \ref{tab:ablation-analysis-e2e} shows the performances of the three systems, we can find that:

1) One pass paradigm for E$^{3}$C can effectively share and reinforce the decisions between two tasks.
Compared with $\text{E}^{3}\text{C}_\text{Two Stage}$, which has a comparable event detection performance (Type-F1), $\text{E}^{3}\text{C}$ gains 5.2\% AVG-F on the downstream event coreference task.

2) Incorporating mention proposal loss can significantly enhance mention detection performance.
By removing mention proposal loss, $\text{E}^{3}\text{C}$ will loss 2.3\% and 4.2\% on Type-F1 and AVG-F, respectively.
Additionally, the coreference performance can be further significantly improved if golden mentions are used -- from 40.89 $\text{E}^{3}\text{C}$ to 53.72 of $\text{E}^{3}\text{C}_\text{GoldMention}$.
This shows that event detection is still a bottleneck for event coreference.

\paragraph{Effect of Pre-trained Models.}

Pre-trained models are important for neural network-based methods.
To investigate their effect on $\text{E}^{3}\text{C}$, Table \ref{tab:ablation-analysis-pretrain-model} shows the performance  of $\text{E}^{3}\text{C}$ using ELMo \citep{peters-etal:2018:naacl2018}, $\text{BERT}_{\text{BASE-Cased}}$, $\text{BERT}_{\text{LARGE-Uncased}}$, $\text{BERT}_{\text{LARGE-WWM-Uncased}}$ \citep{devlin-etal:2019:bert}, GloVe \citep{pennington-etal:2014:glove} 300-dimensional word embedding and char embeddings where the contextual layer is BiLSTM. We can find that:

1) 
Due to the diversity of event mentions, pre-trained contextualized embeddings are critical for mention representation.
All contextualized embeddings outperform GloVe by a large margin in both event detection and event coreference.

\begin{table}[!bht]
  \centering
  \resizebox{0.32\textwidth}{!}{
  \begin{tabular}{lcc}
    \toprule
    & \textbf{AVG-F} & $\mathbf{\Delta}$\\
    \midrule
    E$^3$C & 40.85 \\
    ~~ - Type-Refined & 39.61 & -1.24\\
    ~~ - Type-Guided & 37.80 & -3.05\\
    % \midrule
    Type Rule & 31.68 & -9.17\\
    \bottomrule
  \end{tabular}%
  }
  \caption{
    Ablation results of type-guided mechanism on KBP 2017.
    }
  \label{tab:ablation-analysis-type}%
\end{table}%

\begin{table}[!bht]
  \centering
  \resizebox{0.36\textwidth}{!}{
  \begin{tabular}{lcc}
    \toprule
    & \textbf{Type-F1}   & \textbf{AVG-F}\\
    \midrule
    E$^3$C & 58.33 & 40.85 \\
    % \midrule
    E$^3$C$_{\text{Two Stage}}$ & 57.63 & 38.82\\
    E$^3$C$_{\text{w/o Proposal Loss}}$ & 56.98 & 39.14\\
    % \midrule
    E$^3$C$_{\text{GoldMention}}$ & 72.73 & 53.72\\
    \bottomrule
  \end{tabular}%
  }
  \caption{Performance of different $\text{E}^{3}\text{C}$ settings on KBP 2017.}
  \label{tab:ablation-analysis-e2e}%
\end{table}

\begin{table}[!bht]
  \setlength{\belowcaptionskip}{-0.25cm}
    \centering
    \resizebox{0.48\textwidth}{!}{
    \begin{tabular}{lcc}
      \toprule
      $\textbf{E}^{3}\textbf{C}$ & \textbf{Type-F1}   & \textbf{AVG-F}\\
      \midrule
      \, $\text{BERT}_{\text{BASE-Uncased}}$ (\textit{this paper}) & 58.33 & 40.85\\
      \midrule
      \, GloVe + Char + BiLSTM & 52.45 & 36.43 \\
      \, ELMo & 55.24 & 37.27\\
      \, $\text{BERT}_{\text{BASE-Cased}}$ & 57.08 & 39.14\\ 
      \, $\text{BERT}_{\text{LARGE-Uncased}}$ & 58.05 & 40.99\\
      \, $\text{BERT}_{\text{LARGE-WWM-Uncased}}$ & \textbf{59.29} & \textbf{42.23}\\
      \bottomrule
    \end{tabular}%
    }
    \caption{Performance using different pretrained models for $\text{E}^{3}\text{C}$ on KBP 2017.}
    \label{tab:ablation-analysis-pretrain-model}%
  \end{table}%

\begin{table}[htbp]
    \centering
    \begin{tabular}{lccc}
      \toprule
      & \textbf{Type-F1}   & \textbf{AVG-F}\\
      % \hline
      \midrule
      NW & 59.27 & 42.39\\
      DF & 57.38 & 39.28\\
      \bottomrule
    \end{tabular}%
    % }
    \caption{Results on subsets of different genres in KBP 2017.
    NW indicates newswire documents, while DF indicates discussion forum threads.
    }
    \label{tab:ablation-analysis-domain}%
\end{table}%

2)
$\text{E}^{3}\text{C}$ can be further improved by employing better pre-trained contextual embeddings.
Compared with $\text{BERT}_{\text{BASE-Uncased}}$ used in this paper, $\text{E}^{3}\text{C}$ equipped with $\text{BERT}_{\text{LARGE-WWM-Uncased}}$ gains 1.6\% Type-F1 and 3.4\% AVG-F improvements.

\subsection{Discussions} \label{sec:discussions_and_challenges}

\paragraph{Event Detection Bottleneck.}
From the above experiments, we find that one main bottleneck of event coreference is event detection.
As shown in Table \ref{tab:ablation-analysis-e2e}, using gold mentions results in a massive improvement on AVG-F, from 40.85 to 53.72.
Besides, even if we fix all coreference link errors in predicted event detection results, the growth of AVG-F is still limited, from 40.85 to 42.80.
Event detection is challenging because:
1) Event mentions are diversified and ambiguous, detecting them requires a deep understanding of contexts.
2) Some event mentions are multi-tagged\footnote{10.18\% in KBP 2016 and 8.4\% in KBP 2017}, i.e., one span triggering multiple events.
Because this paper does not consider this issue, it misses some mentions.

\paragraph{Domain Adaptation.}
We find that domain adaptation is another challenge for event coreference.
Table \ref{tab:ablation-analysis-domain} shows the results of our $\text{E}^{3}\text{C}$ model on different genres of KBP 2017 evaluation dataset: 83 newswire documents -- NW, and 84 discussion forum threads -- DF.
There is a significant performance gap between the two genres, probably because:
1) Different from formal NW documents, DF threads are often informal and lack coherent discourse structures \citep{choubey-huang:2018:acl2018}.
2) Event chains in a discussion forum thread are not only relevant to contents, but also to speaker information and discussion topic. Solving this problem requires a deep understanding of dialogue contexts.

\paragraph{Argument Modeling.}
In this paper, we exploited the argument’s information implicitly via a mask attention strategy, without explicitly extracting argument role.
However, we believe event coreference can be further enhanced by modeling argument information more effectively:
1) incorporating explicit argument information can effectively capture semantic information of events for better feature representation \cite{peng-etal-2016-event,choubey-huang:2017:emnlp2017};
2) the coreference/compatibility of argument is crucial for deciding coreference relations between events \cite{Lee:2012:JEE:2390948.2391006,huang-etal:2019:naacl2019}.
Unfortunately, the traditional argument-based end-to-end pipeline event coreference methods \cite{chen-ng:2014:LREC2014,yang-etal-2015-hierarchical} suffer from the error propagation problem of previous components, e.g., argument extraction and entity coreference.
The denoising feature composition algorithms or joint modeling of entity/event coreference may effectively solve the argument's error propagation problem.

\section{Related Work} \label{sec:related_work}
\paragraph{Event Coreference.}
Event coreference aims to cluster textual mentions of the same event.
Different from cross-document event coreference works \citep{yang-etal-2015-hierarchical,zhang-etal-2015-cross,choubey-huang:2017:emnlp2017,kenyon-dean-etal-2018-resolving,barhom-etal-2019-revisiting}, this paper focuses on the within-document event coreference task.

Traditional approaches \citep{chen-ji:2009:graph2009,chen-ng:2014:LREC2014,liu-etal:2014:lrec2014} are mostly pipeline-based systems depending on several upstream components, thus often suffer from the error propagation problem.
To address this problem, many joint models have been proposed, e.g., joint inference \citep{Chen:2016:AAAI2016:JointInference,lu-etal-joint:2016:COLING2016} and joint modeling \citep{araki-mitamura:2015:EMNLP2015:joint,lu-ng-joint:2017:acl2017}.
Furthermore, the above methods use hand-crafted features, which are hard to generalize to the new languages/domains/datasets.
Several neural network models \citep{krause-etal-2016-event,chao:2019:selective} and transfer techniques \citep{huang-etal:2019:naacl2019} are proposed to complement these methods with automatic feature learning abilities.

Compared to previous approaches, $\text{E}^{3}\text{C}$ is the first fully end-to-end neural event coreference resolution approach.
It can extract features, detect event mentions, and resolve event chains in the same network.

\paragraph{End-to-end Entity Coreference.}
Recently, end-to-end neural networks \citep{lee-etal:2017:emnlp2017,lee-etal:2018:naacl2018,kantor-globerson:2019:acl2019,fei-etal-2019-end,joshi-etal:2019:emnlp2019:bertforcoref} have achieved significant progress in entity coreference.
These methods also motivate this study.
Due to the mention diversity and the long-distance coreference problems, event coreference is usually considered more challenging than entity coreference \citep{lu-ng:2018:ijcai2018,choubey-huang:2018:acl2018}.
This paper proposes a type-guided mechanism, where can resolve the above challenges by incorporating type information, learning semantic event mention representation, and modeling long-distance, semantic-dependent evidence.

\section{Conclusions} \label{sec:conclusion}

This paper proposes a state-of-the-art, end-to-end neural network for event coreference resolution -- $\text{E}^{3}\text{C}$ neural network, which jointly models event detection and event coreference, and learns to extract features from the raw text directly.
A type-guided mechanism is further proposed for resolving the mention diversity problem and the long-distance coreference problem, which: 1) informs coreference prediction with type scoring, 2) refines mention representation using type information, and 3) guides decoding under type consistency.
Experiments show that our method achieves state-of-the-art performances on KBP 2016 and KBP 2017.
For future work, we will focus on the bottleneck of event coreference, e.g., event detection and argument modeling.

\bibliography{e2e_event_coref}
\bibliographystyle{acl_natbib}

\appendix
\section{Experiment Details}
Table \ref{tab:hyper-parameters} presents the detailed hyper-parameters of the E$^{3}$C model used in our experiments.
And we conducted all experiments on a Nvidia TITAN RTX GPU.
\begin{table}[!htb]
  \centering
  \resizebox{0.48\textwidth}{!}{
  \begin{tabular}{lr}
    \toprule
   Parameter name & Parameter value\\
    \midrule
    Mini batch size & 1 \\
    Max epochs for stopping training & 150 \\
    Patience for early stopping & 10 \\
    Max antecedents number & 50 \\
    Max document length for training & 1024 \\
    Dropout for word representation & 0.5 \\
    Dropout for FFNN & 0.2 \\
    Hidden layers for FFNN & 2 \\
    Hidden units for FFNN & 150 \\
    Optimizer & Adamax \\
    Initial learning rate & 0.001 \\
    Learning rate anneal factor & 0.5 \\
    Learning rate anneal patience & 5 \\
    \bottomrule
  \end{tabular}%
  }
  \caption{Hyper-parameters of the E$^{3}$C model used in our experiments. FFNN indicates the feed-forward neural networks for mention proposaling and antecedent scoring.
  }
  \label{tab:hyper-parameters}%
\end{table}%

\section{Data Sets}

We used Stanford CoreNLP toolkit\footnote{https://stanfordnlp.github.io/CoreNLP/} to preprocess all documents for xml tags cleaning, sentence splitting and tokenization.
Since only 18 categories were used for the official evaluation, we filtered out event instances with other categories in the training data.

\section{Reproducibility}
In this section, we present the reproducibility information of the paper.
Table \ref{tab:validation_performance} shows the corresponding validation performance for all reported KBP 2016 and KBP 2017 results.
In addition, Table \ref{tab:runtime_num_param} presents the average runtime for each approach and number of parameters in each model.

\begin{table*}[htbp]
  \centering
  \resizebox{0.98\textwidth}{!}{
    \begin{tabular}{clcccccccc}
      \toprule
          & & \textbf{Type-P} & \textbf{Type-R} & \textbf{Type-F1} & \textbf{B}$_{3}$    & \textbf{CEAF}$_{e}$ & \textbf{MUC}   & \textbf{BLANC} & \textbf{AVG-F} \\
          \midrule
          \multirow{2}{*}{KBP 2016} &  Interact Model$_{\text{BERT}}$ & 56.8 & 59.18 & 57.97 & 45.49 & 44.07 & 37.43 & 30.41 & 39.35 \\
          & E$^3$C & 62.94 & 59.10 & 60.96 & 49.02 & 46.76 & 42.80 & 33.00 & 42.89  \\
          \midrule
          \multirow{13}{*}{KBP 2017} & Interact Model$_{\text{BERT}}$ & 60.97 & 57.36 & 59.11 & 49.72 & 50.30 & 32.50 & 32.06 & 41.14 \\
          & E$^3$C & 65.85  & 54.61  & 59.71  & 51.60  & 51.48  & 39.45  & 35.01  & 44.38  \\
          & E$^3$C$_{\text{w/o Type-Refined}}$ & 70.33 & 48.64 & 57.51 & 50.45 & 49.96 & 40.15 & 34.16 & 43.68   \\
          & E$^3$C$_{\text{w/o Type-Guided}}$ & 65.97  & 54.93  & 59.94  & 47.26  & 41.93  & 37.22  & 33.54  & 39.99  \\
          & E$^3$C$_{\text{Two Stage}}$ & 63.92  & 55.26  & 59.27  & 51.08  & 49.80  & 35.71  & 33.91  & 42.62  \\
          & E$^3$C$_{\text{w/o Proposal Loss}}$ & 68.15  & 51.33  & 58.56  & 50.80  & 51.29  & 37.05  & 34.14  & 43.32  \\
          & E$^3$C$_\text{GloVe+Char}$ & 63.92  & 50.50  & 56.43  & 47.86  & 45.76  & 36.45  & 31.92  & 40.50  \\
          & E$^3$C$_\text{ELMo}$ & 64.42  & 53.74  & 58.60  & 50.57  & 50.70  & 35.25  & 33.42  & 42.48  \\
          & E$^3$C$_\text{BERT-BASE-Cased}$ & 68.08  & 52.75  & 59.45  & 51.46  & 50.53  & 38.66  & 34.85  & 43.88  \\
          & E$^3$C$_\text{BERT-LARGE-Uncased}$ & 66.65 & 55.97 & 60.85 & 53.21 & 52.60 & 41.69 & 37.37 & 46.22   \\
          & E$^3$C$_\text{BERT-LARGE-WWM-Uncased}$ & 69.48  & 55.01  & 61.40  & 53.78  & 53.01  & 41.62  & 36.80  & 46.30  \\
          \cmidrule{2-10}
          & Type-Rule & 65.85  & 54.61  & 59.71  & 35.88  & 26.91  & 29.39  & 25.91  & 29.52  \\
          & E$^3$C$_{\text{GoldMention}}$ & 81.39  & 70.95  & 75.81  & 67.76  & 66.41  & 48.16  & 51.60  & 58.48  \\
    \bottomrule
    \end{tabular}%
  }
  \caption{Corresponding validation performance for each reported KBP 2016/2017 result.
  Type-Rule and E$^3$C$_{\text{GoldMention}}$ take unreal experiment setups for exploiting the bound performance of E$^3$C.
  Type Rule is a simple heuristic method that regards all event mentions in the same type are coreferential, and it directly uses the event detection result from E$^{3}$C.
  E$^3$C$_{\text{GoldMention}}$ uses gold mentions instead of mentions proposed by the mention proposal layer, but the model still needs to predict the type of each mention.
  }
  \label{tab:validation_performance}%
\end{table*}%

\begin{table*}[htbp]
  \centering
  \resizebox{0.60\textwidth}{!}{
    \begin{tabular}{lrc}
      \toprule
          & \multicolumn{1}{c}{Time for one epoch (s)} & \multicolumn{1}{c}{$|\mathbf{\Theta}_{\text{update}}|$}  \\
    \midrule
    E$^3$C & 82.76  & 2,886,108 \\
    Interact Model$_\text{BERT}$ & 408.25  & 3,613,431 \\
    E$^3$C$_{\text{w/o Type-Refined}}$ & 78.77  & 1,705,692 \\
    E$^3$C$_{\text{w/o Type-Guided}}$ & 81.80  & 2,886,108 \\
    E$^3$C$_{\text{Two Stage}}$ & 75.52  & 1,708,561 \\
    E$^3$C$_{\text{w/o Proposal Loss}}$ & 80.33  & 2,886,108 \\
    E$^3$C$_\text{GloVe + Char}$ & 100.00  & 2,272,595 \\
    E$^3$C$_\text{ELMo}$ & 1345.30  & 4,880,339 \\
    E$^3$C$_\text{BERT-BASE-Cased}$ & 82.74  & 2,886,108 \\
    E$^3$C$_\text{BERT-LARGE-Uncased}$ & 152.23  & 4,880,360 \\
    E$^3$C$_\text{BERT-LARGE-WWM-Uncased}$ & 152.47  & 4,880,360 \\
    \bottomrule
    \end{tabular}%
  }
    \caption{Average runtime for each approach and number of parameters in each model.
    $\mathbf{\Theta}_{\text{update}}$ refers to the number of trainable parameters.
    We fix all parameters for word representations in our experiments, such as BERT, GloVe, and ELMo parameters.}
  \label{tab:runtime_num_param}%
\end{table*}%

\end{document}